\documentclass[10pt,twocolumn,letterpaper]{article}

\usepackage{iccv}
\usepackage{times}
\usepackage{epsfig}
\usepackage{graphicx}
\usepackage{amsmath}
\usepackage{amssymb}
\usepackage{bm}
\usepackage{subcaption}
\usepackage[numbers,sort]{natbib} 
\usepackage{booktabs}
\usepackage{multirow}

\usepackage[pagebackref=true,breaklinks=true,letterpaper=true,colorlinks,bookmarks=false]{hyperref}

\newcommand{\ifArXiv}[2]{#1}  

\def\C2{$\texttt{C}^\texttt{2}$\xspace}

\DeclareMathOperator*{\argmin}{arg\,min}

\hypersetup{
	plainpages=false,
	colorlinks=true,              %
	linkcolor=blue,               %
	anchorcolor=blue,             %
	citecolor=blue,               %
	filecolor=blue,               %
	urlcolor=blue,                %
	pdfview=FitH,                 %
	pdfstartview=FitH,            %
	pdfpagelayout=SinglePage      %
}

\iccvfinalcopy 

\def\iccvPaperID{4034} 

\ificcvfinal\fi
\begin{document}

\title{Controllable Attention for Structured Layered Video Decomposition}

\author{Jean-Baptiste Alayrac\textsuperscript{1}\thanks{Equal contribution.}
	\and
	Jo\~ao Carreira\textsuperscript{1}\footnotemark[1]
	\and
	Relja Arandjelovi\'c\textsuperscript{1}
	\and
	Andrew Zisserman\textsuperscript{1,2}
	\and 
	{\tt\small \{jalayrac,joaoluis\}@google.com}	
	\\
	\textsuperscript{1}DeepMind
	\quad
	\textsuperscript{2}VGG, Dept.\  of Engineering Science, University of Oxford
}

\maketitle

\begin{abstract}
The objective of this paper is to be able to separate a video into its natural layers, 
and to control which of the separated layers to attend to.
For example, to be able to separate reflections,  transparency or object motion.

We make the following three contributions:
(i)  we introduce
a new 
structured neural network architecture that explicitly incorporates layers (as spatial masks) into its design.
This improves separation performance over previous general purpose networks for this task;
(ii)  we
demonstrate that we can augment the architecture to leverage external
cues such as \textit{audio} for controllability and to help disambiguation; and (iii) we experimentally
demonstrate the effectiveness of our approach and training procedure
with controlled experiments while also showing that the proposed model
can be successfully applied to real-word applications
such as \textit{reflection removal} and \textit{action recognition in
cluttered scenes}.
\end{abstract}

\section{Introduction}
\label{sec:intro}

``The more you look the more you see", is generally true for our
complex, ambiguous visual world. Consider the everyday task of
cleaning teeth in front of a mirror. People performing this task may
first attend to the mirror surface to identify any dirty spots, clean
them up, then switch attention to their mouth
reflected in the mirror. Or they may hear steps
behind them and switch attention to a new face now reflecting in the
mirror. Not all visual possibilities can be investigated at once given
a fixed computational budget and this creates the need for such
controllable attention mechanisms.

\begin{figure}[t]
	\begin{center}
		\includegraphics[width=\linewidth]{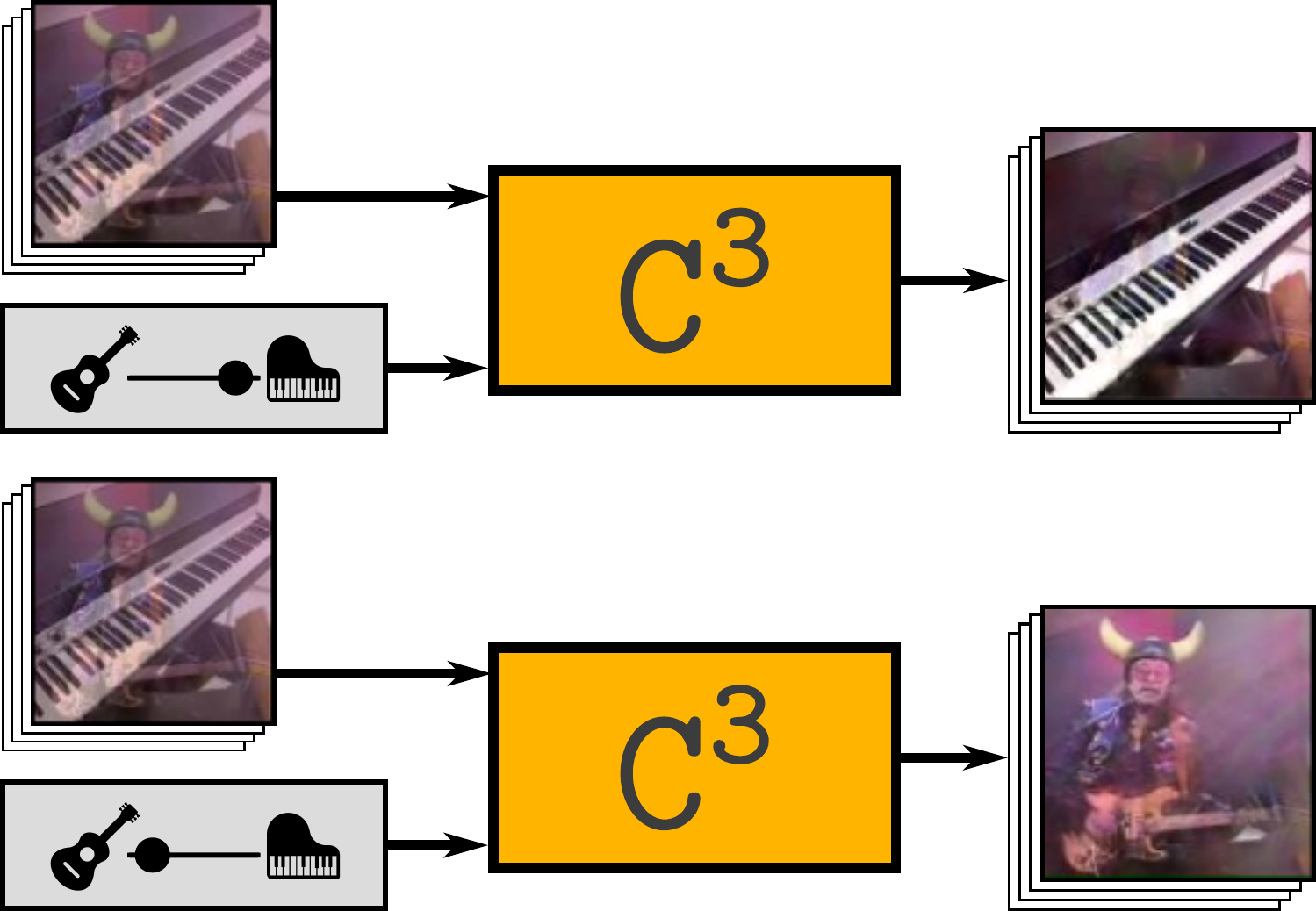}
	\end{center}
	\caption{\label{fig:teaser} \small 
		We propose a model, $\texttt{C}^\texttt{3}$, able to decompose a video into meaningful layers. 
		This decomposition process is controllable through external cues such as \emph{audio}, that can select the layer to output.}
\end{figure}

Layers offer a simple but useful model for handling this complexity of the visual 
world~\cite{Wang1994RepresentingMI}. 
They provide a compositional
model of an image or video sequence, and cover a multitude of scenarios (reflections, shadows, occlusions,
haze, blur, ...) according to the composition rule. For example, an additive composition models reflections,
and occlusion is modelled by superimposing opaque layers in a depth ordering.
Given a a layered decomposition, attention can switch between the various layers as necessary for the task at hand.

Our objective in this paper is to separate videos into their
constituent layers, and to {\em select} the layers to attend to as illustrated in Figure~\ref{fig:teaser}.  A
number of recent works have used deep learning to separate layers in
images and videos~\cite{Fan2017,Heydecker2018,Chi2018,
Zhang2018SingleIR,alayrac19centrifuge,gandelsman2018double}, with varying success, but the selection of the
layers has either had to be hard coded into the architecture, or the
layers are arbitrarily mapped to the outputs. For example,
\cite{alayrac19centrifuge} considers the problem of separating blended videos
into component videos, but because the mapping between input videos
and outputs is arbitrary,
training is forced to use a permutation
invariant loss, and there is no control over the mapping at
inference time.
How can this symmetry between the composed input
layers and output layers be broken? 

The solution explored here is based on the simple fact  that videos do not consist of visual streams alone, they
also have an audio stream;  and,  significantly, the visual and audio streams are often correlated.
The correlation can be strong
(e.g.\ the synchronised sound and movement of beating on a drum), or quite weak (e.g.\ street noise that separates
an outdoor from indoor scene), but this correlation can be employed to break the symmetry. 
This symmetry breaking is related  to recent approaches to the 
cocktail party audio separation problem~\cite{Afouras2018, Ephrat2018} where  visual cues are used to
select speakers and improve the quality of the separation. Here we use audio cues to select the visual layers.

\vspace{3mm}
\noindent
\textbf{Contributions:} 
The contributions of this paper are threefold: \textbf{(i)} we propose
a new 
structured neural network architecture that explicitly incorporates layers (as spatial masks) into its design;
 \textbf{(ii)} we
demonstrate that we can augment the architecture to leverage external
cues such as \textit{audio} for controllability and to help disambiguation; and \textbf{(iii)} we experimentally
demonstrate the effectiveness of our approach and training procedure
with controlled experiments while also showing that the proposed model
can be successfully applied to real-word applications
such as \textit{reflection removal} and \textit{action recognition in
cluttered scenes}.

We show that the new architecture leads to improved layer separation. This is
demonstrated both qualitatively and quantitatively by comparing to
recent general purpose models, such as the visual centrifuge~\cite{alayrac19centrifuge}. For the
quantitative evaluation we evaluate how the downstream task of human
action recognition is affected by reflection removal. For this, we
compare the performance of a standard action classification network on
sequences with reflections, and with reflections removed using the
layer architecture, and demonstrate a significant improvement in the
latter case.

\section{Related work}
\label{sec:rw}

\vspace{3mm}
\noindent \textbf{Attention control.} 
Attention in neural network modelling has had  a significant impact in natural language processing, such as machine translation,
\cite{bahdanau2014neural,vaswani2017attention} and
vision~\cite{xu2015show}, where it is implemented as a soft masking of
features. In these settings attention is often not directly evaluated,
but is just used as an aid to improve the end performance. In this
paper we investigate models of attention in isolation, aiming for high
consistency and controllability. By consistency we mean the ability to
maintain the focus of attention on a particular target. By
controllability we mean the ability to switch to a different target on
command.

Visual attentional control is actively studied in psychology and
neuroscience~\cite{yantis1998control,egeth1997visual,gazzaniga2013cognitive,oliva2003top,Tsotsos95,Itti05}
and, when malfunctioning, is a potentially important cause of
conditions such as ADHD, autism or
schizophrenia~\cite{mash2012abnormal}. One of the problems studied in
these fields is the relationship between attention control based on
top-down processes that are voluntary and goal-directed, and bottom-up
processes that are stimulus-driven (e.g.\  saliency)
\cite{itti1998model,Tsotsos95}. Another interesting aspect is the types of
representations that are subject to attention, often categorized into
location-based~\cite{siegel2008neuronal}, object-based or
feature-based~\cite{baldauf2014neural}: examples of the latter include
attending to anything that is red, or to anything that moves. 
Another
relevant stream of research relates to the role of attention in
multisensory
integration~\cite{spence1996audiovisual,talsma2010multifaceted}. Note
also that attention does not always require eye movement -- this is
called \textit{covert} (as opposed to \textit{overt}) attention. In
this paper we consider covert attention as we will not be considering
active vision approaches, and focus on feature-based visual attention
control.

\vspace{3mm}
\noindent \textbf{Cross-modal attention control.}
The idea of using one modality to control attention in the other has
a long history, one notable application being informed audio source separation and denoising \cite{Girin01,Wang05,Barzelay07,Segev12}.
Visual information has been used to aid audio denoising \cite{Girin01,Segev12},
solve the cocktail party problem of isolating sound coming from
different speakers \cite{Wang05,Afouras2018,Ephrat2018,Owens18} or
musical instruments \cite{Barzelay07,Gao18,Zhao18}.
Other sources of information used for audio source separation include
text to separate speech \cite{LeMagoarou15}
and score to separate musical instruments \cite{Hennequin11}.

More relevant to this paper where audio is used for control,
\cite{Arandjelovic18,Owens18,Senocak18,Zhao18} learn to attend to the object
that is making the sound. However, unlike in this work,
they do not directly output the disentangled video
nor can they be used to remove reflections as objects are assumed to be perfectly
opaque.

Other examples of control across modalities include
temporally localizing a moment in a video using language \cite{Hendricks17},
video summarization guided by titles \cite{Song15} or query object labels \cite{Sharghi16}, object localization from spoken words \cite{Harwath18}, image-text alignment \cite{Karpathy15}, and interactive object segmentation via user clicks \cite{Boykov01}.

\vspace{3mm}
\noindent \textbf{Layered video representations.} 
Layered image and video representations have a long history in computer
vision~\cite{wang1994representing} and are an appealing framework for
modelling 2.1D depth
relationships~\cite{smith2004layered,yang2010layered}, motion
segmentation~\cite{wang1994representing},
reflections~\cite{Farid1999,Fan2017,Heydecker2018,Chi2018,Guo2014,Beery2008,
KunGai2012, Szeliski2000, Xue2015, Nandoriya2017,Zhang2018SingleIR},
transparency~\cite{alayrac19centrifuge,gandelsman2018double}, or even
haze~\cite{gandelsman2018double}. There is also evidence that the
brain uses multi-layered visual representations for modelling
transparency and occlusion~\cite{webster2009color}.

\begin{figure*}[t!]
	\centering
	\begin{subfigure}[t]{.53\linewidth}
		\centering
		\includegraphics[height=3.8cm]{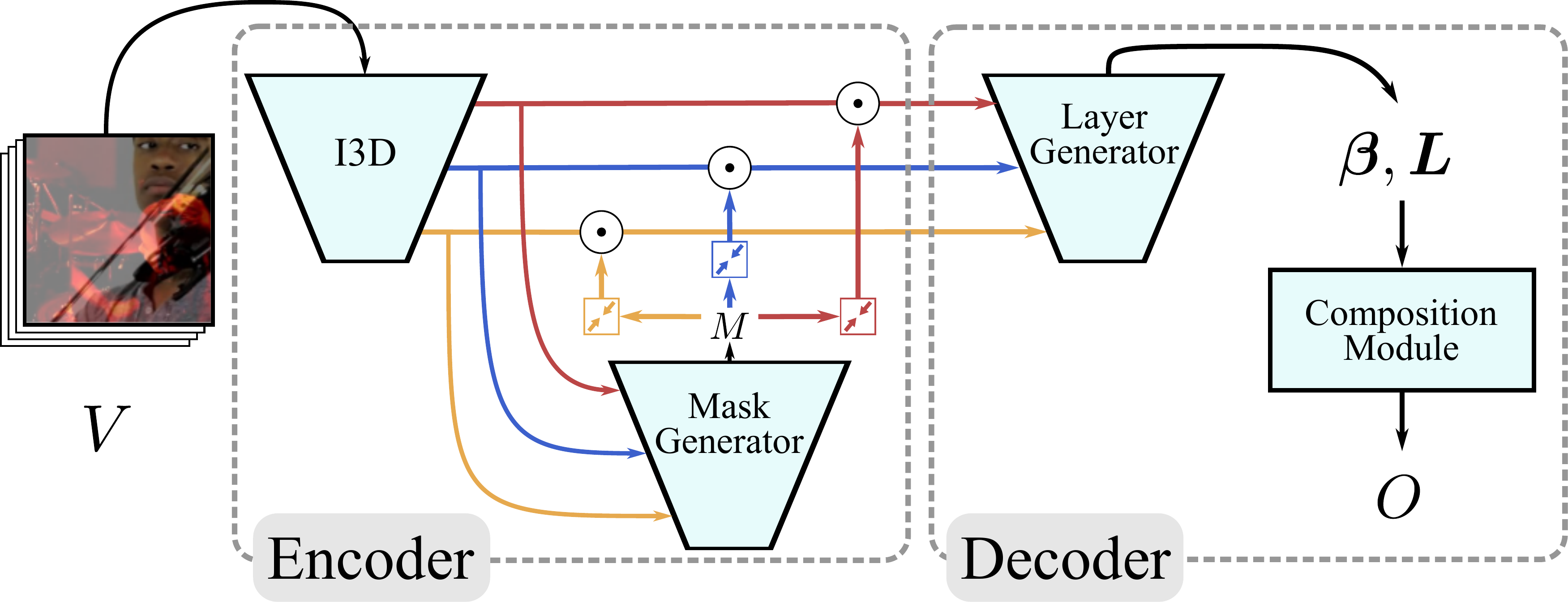} 
		\caption{\small\textbf{Overview of the Compositional Centrifuge ($\texttt{C}^\texttt{2}$) architecture} \label{fig:architectures}}
	\end{subfigure}
	\hfill
	\begin{subfigure}[t]{.44\linewidth}
		\centering
		\includegraphics[height=3.8cm]{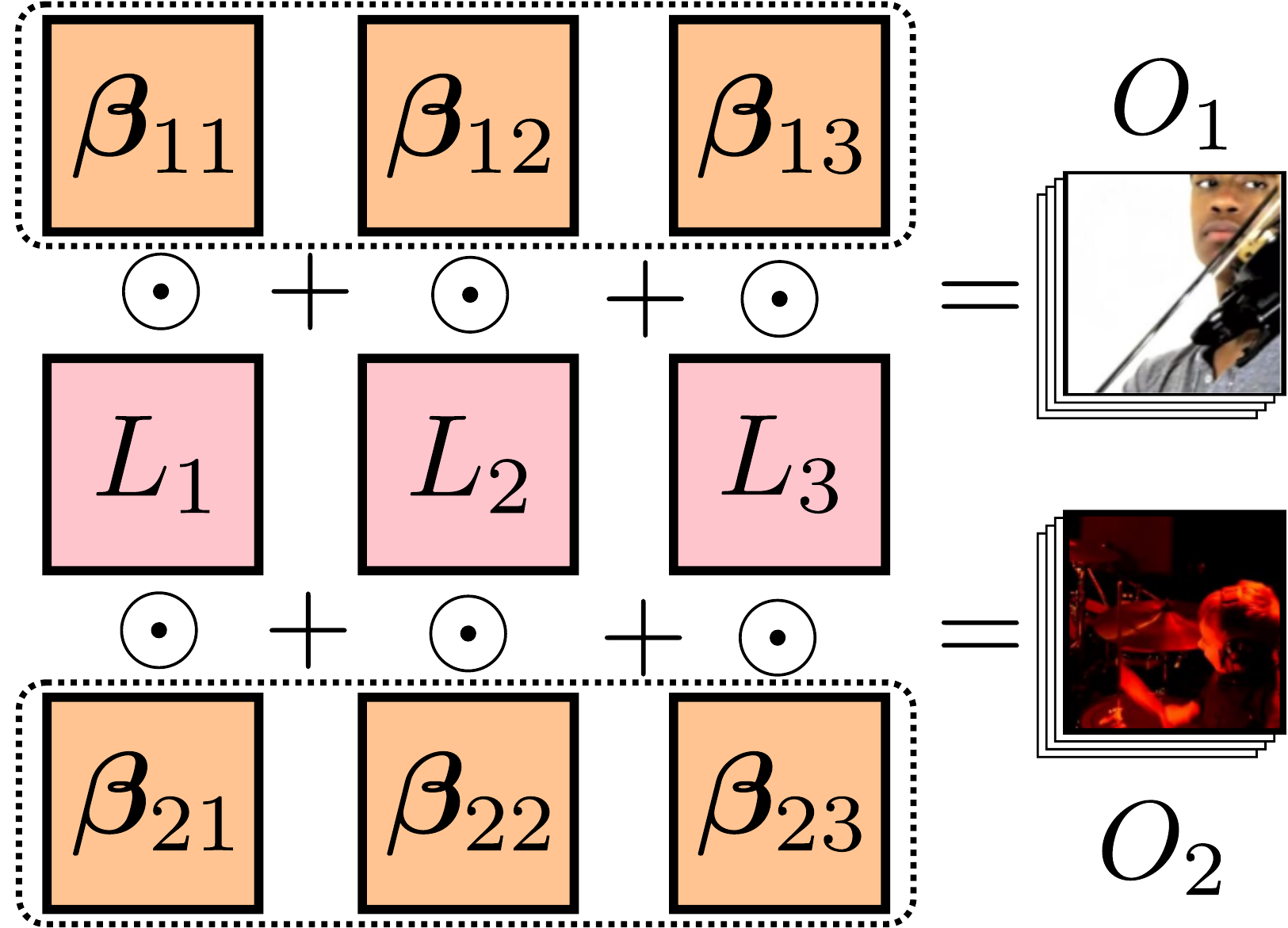} 
		\caption{\small \textbf{Composition module}  \label{fig:architectures_comp}}
	\end{subfigure}
	\vspace*{-0.1cm}
	\caption{\label{fig:model} \textbf{Network architecture for layer decomposition} (\ref{sec:archi}).
	}	
	\vspace{-0.4cm}
\end{figure*}

\section{Approach}
\label{sec:method}

This section describes the two technical contributions of this work.
First, in Section~\ref{sec:archi}, a novel architecture for
decomposing videos into layers.  This architecture is built upon the
visual centrifuge~\cite{alayrac19centrifuge}, a generic U-Net like
encoder-decoder, but extends it with two structural changes tailored towards the layered video decomposition task.
Second, in Section~\ref{sec:external},  the decomposition model is endowed with
controllability -- the ability of the network to use
external cues to control what it should focus on reconstructing.  Here, we propose to use a natural video modality,
namely \emph{audio}, to select layers.  Given this external cue,
different mechanisms for controlling the outputs are investigated.
Finally, in Section~\ref{sec:datagen}, we describe how this model can be trained for successful
controllable video decomposition.

In the following, $V$ stands for an input video.
Formally, $V\in\mathbb{R}^{T\times W \times H \times 3}$ where $T$ is the number of frames, $W$ and $H$ are the width and height of the frames, and there are $3$ standard RGB channels.
The network produces an $T\times W \times H \times (n \times 3)$ tensor,
interpreted as $n$ output videos $\bm{O}$, where each $O_i$ is of the same size as $V$.

\subsection{Architecture for layer decomposition}
\label{sec:archi} 

We start from the visual centrifuge~\cite{alayrac19centrifuge}, a
U-Net~\cite{Ronneberger2015} encoder-decoder architecture, which
separates an input video into $n$ output videos.  The encoder consists
of an I3D network~\cite{carreira17quovadis} and the decoder is
composed by stacking 3D up convolutions.  However, the U-Net
architecture used there is generic and not tailored to the layered
video decomposition task (this is verified experimentally in
Section~\ref{subsec:quant}).  Therefore, we propose two structural
modifications specifically designed to achieve layered decomposition,
forming a new network architecture, \emph{Compositional Centrifuge
($\texttt{C}^\texttt{2}$)}, shown in Figure~\ref{fig:architectures}.
Firstly, a bespoke gating mechanism is used in the encoder, which
enables selection of scene segments across space/time, thereby
making the decoder's task easier.  Secondly, layer compositionality is
imposed by constraining how the output videos are generated -- the
\emph{layer generator} outputs multiple layers $\bm{L}$ and their
composing coefficients $\bm{\beta}$ such that the output videos
$\bm{O}$ are produced as a linear combination of the layers.  These
modifications are described in detail next.

\vspace{3mm}
\noindent
\textbf{Encoder.} 
We aim to recover layers in the presence of occlusions and transparent
surfaces. In such cases there are windows of opportunity when objects
are fully visible and their appearance can be modelled, and periods
when the objects are temporarily invisible or indistinguishable and
hence can only be tracked. We incorporate this intuition into a novel
spatio-temporal encoder architecture. The core
idea is that the features produced by the I3D are gated with
multiple ($m$) masks, also produced by the encoder itself.
The gated features therefore already encode information about the underlying layers
and this helps the  decoder's task.

In order to avoid gating all features with all $m$ masks, which would
be prohibitively expensive in terms of computation and memory usage,
feature channels are divided into $m$ mutually-exclusive groups
and each mask is applied only to the corresponding group.

More formally, the mask generator produces $\bm{M}\in\left[ 0,1\right]^{T\times W \times H \times m}$ which is interpreted as a set of $m$ spatio-temporal masks $\bm{M}=(M^c)_{c=1}^m$.
$\bm{M}$ is constrained to sum to $1$ along the channel dimension by using a softmax nonlinearity.
Denote $F_l$ the output feature taken at level $l$ in the I3D.
We assume that $F_l\in\mathbb{R}^{T_l\times W_l \times H_l \times (m\times d_l)}$, \ie the number of output channels of $F_l$ is a multiple of $m$.
Given this, $F_l$ can be grouped into $m$ features $(F_l^c)_{c=1}^m$ where $F_l^c\in\mathbb{R}^{T_l\times W_l \times H_l \times d_l}$.
The following transformation is applied to each $F_l^c$:

\begin{equation}
\label{eq:gating_masking}
\tilde{F}_l^c=M_l^c\odot F_l^c,
\end{equation}
where $M_l^c$ is obtained by downsampling $M^c$ to the shape $[T_l\times W_l \times H_l]$,
$\odot$ refers to the Hadamard matrix product with a slight abuse of notation as the channel dimension is 
broadcast, \ie the same mask is used across the channels.
This process is illustrated in Figure~\ref{fig:architectures}.
\ifArXiv{Appendix~\ref{app:architecture}}{The extended version of this paper~\cite{alayrac19c3}}
gives details on which feature levels are used in practice.

\vspace{3mm}
\noindent
\textbf{Imposing compositionality.}
In order to bias the decoder towards constructing layered decompositions,
we split it into two parts -- the \emph{layer generator} produces $m$
layers $\bm{L}$ and composing coefficients $\bm{\beta}$ which are 
then combined by the \emph{composition module} to form the final $n$ output videos $\bm{O}$.
The motivation is that
individual layers should ideally represent independent scene units, such as moving objects, reflections or shadows, that can be composed in different ways into full scene videos. The proposed model architecture is designed to impose the inductive bias towards this type of compositionality.

More formally, the \emph{layer generator} outputs a set of $m$ \emph{layers} $\bm{L}=(L_j)_{j=1}^m$, where $L_j\in\mathbb{R}^{T\times H\times W \times 3}$,
and a set of $n\times m$ \emph{composing coefficients}
$\bm{\beta}=(\bm{\beta}_{ij})_{(i,j)\in[\![ 1, n ]\!]  \times [\![ 1, m ]\!] }$. 
These are then combined in the \emph{composition module} (Figure~\ref{fig:architectures_comp}) to produce the final output videos $\bm{O}$:
\begin{eqnarray}
\label{eq:compose}
&O_i =& \sum_j \bm{\beta}_{ij}\odot L_j. 
\end{eqnarray}

\subsection{Controllable symmetry breaking}
\label{sec:external}
The method presented in the previous section is inherently symmetric
-- the network is free to assign videos to output slots  in any order.  In this
section, we present a strategy for controllable attention that is able
to break the symmetry by making use of side-information, a
\emph{control signal}, provided as an additional input to the network.
Audio is used as a natural control signal since it is readily
available with the video. In our mirror example from the introduction, hearing speech
indicates the attention should be focused on the person in the mirror,
not the mirror surface itself.  For the rest of this section, audio is
used as the \emph{control signal}, but the proposed approach remains
agnostic to the control signal nature.

Next, we explain how to compute audio features, fuse them with the visual features, and finally, how to obtain the output video which corresponds to the input audio.
The architecture, named \emph{Controllable Compositional Centrifuge ($\texttt{C}^\texttt{3}$)}, is shown in Figure~\ref{fig:c3}.

\vspace{3mm}
\noindent
\textbf{Audio network.}
The audio first needs to be processed before feeding it as a control signal to the video decomposition model.
We follow the strategy employed in~\cite{Arandjelovic18} to process the audio.
Namely, the log spectrogram of the raw audio signal is computed and treated as an image, and a VGG-like network is used to extract the audio features.
The network is trained from scratch jointly with the video decomposition model.

\vspace{3mm}
\noindent
\textbf{Audio-visual fusion.}
To feed the audio signal to the video model, we concatenate audio features to the outputs of the encoder before they get passed to the decoder.
Since visual and audio features have different shapes --
their sampling rates differ and they are 3-D and 4-D tensors for audio and vision, respectively --
they cannot be concatenated naively.
We make the two features compatible by \textbf{(1)} average pooling the audio features over frequency
dimension, \textbf{(2)} sampling audio features in time to match the number of temporal video feature samples, and \textbf{(3)} broadcasting
the audio feature in the spatial dimensions.
After these operations the audio tensor is concatenated with the visual tensor along the channel dimension.
This fusion process is illustrated in Figure~\ref{fig:c3}.
We provide the full details of this architecture in
\ifArXiv{Appendix~\ref{app:architecture}.}{the extended version of this paper~\cite{alayrac19c3}.}

\begin{figure}[t]
	\begin{center}
		\includegraphics[width=\linewidth]{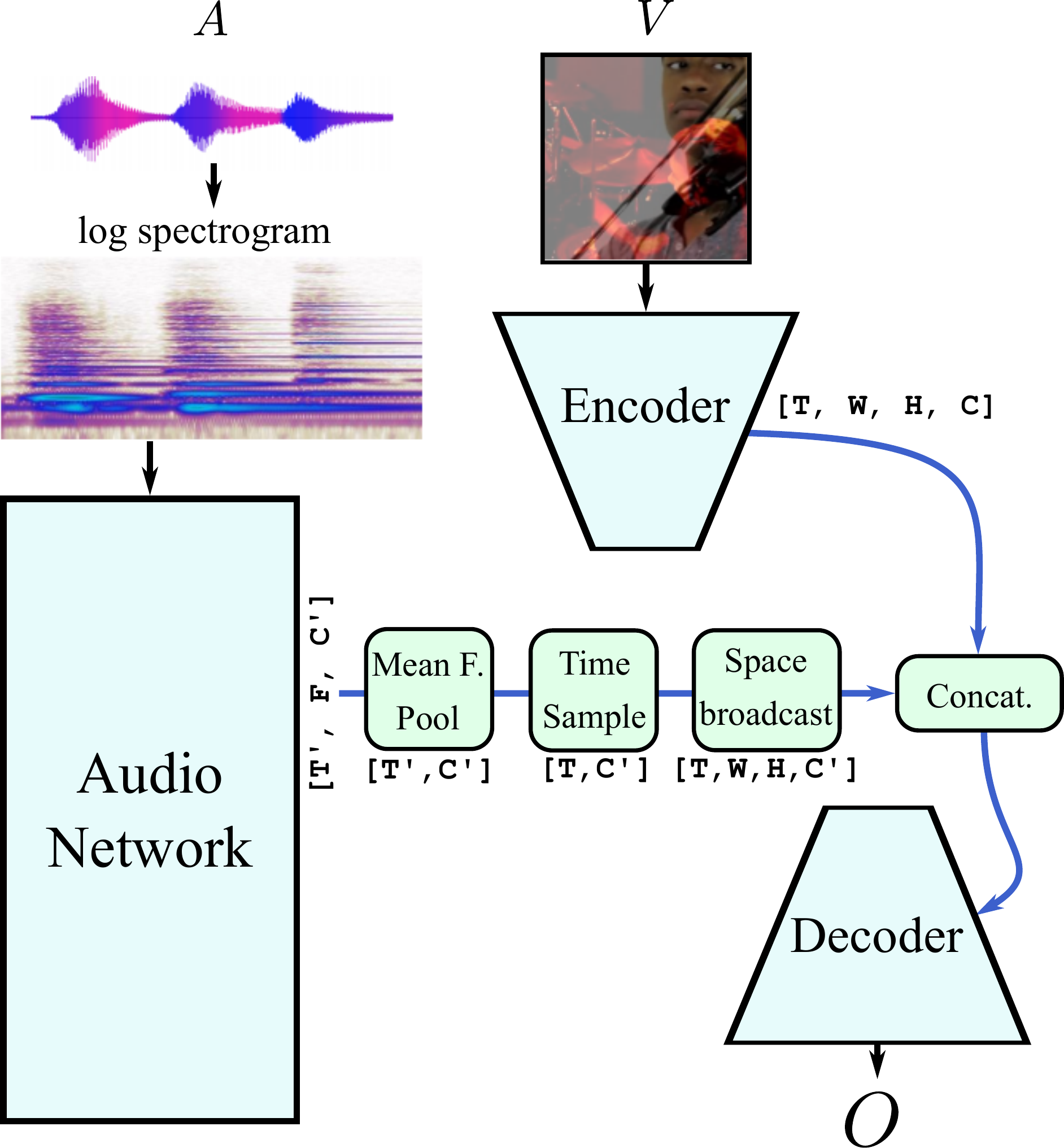}
	\end{center}
	\vspace{-0.4cm}
	\caption{\label{fig:c3} \small
{\bf The Controllable Compositional Centrifuge ($\texttt{C}^\texttt{3}$)}.
The Encoder-Decoder components are the same as in $\texttt{C}^\texttt{2}$ (Figure~\ref{fig:architectures}).
Audio features are extracted from the audio control signal and fused with
the visual features before entering the decoder.}
\vspace{-0.4cm}
\end{figure}

\vspace{3mm}
\noindent
\textbf{Attention control.}
We propose two strategies for obtaining the output video which corresponds to the input audio.
One is to use \emph{deterministic control} where the desired video is forced to
be output in a specific pre-defined output slot, without loss of generality $O_1$ is used.
While simple, this strategy might be too rigid as it imposes too many constraints
onto the network. For example, a network might naturally learn to output guitars in
slot $1$, drums in slot $2$, \etc, while \emph{deterministic control} is forcing it
to change this ordering at will.
This intuition motivates our second strategy -- \emph{internal prediction} --
where the network is free to produce output videos in any order it sees fit,
but it also  provides a pointer to the output slot which contains the desired video.
Internal prediction is trained jointly with the rest of the network, full details of the 
architecture are given in
\ifArXiv{Appendix~\ref{app:architecture}.}{the extended version of this paper~\cite{alayrac19c3}.}
The training procedure and losses for the two control strategies are described in the next section.

\subsection{Training procedure}
\label{sec:datagen}

\vspace{3mm}
\noindent
\textbf{Training data.}
Since it is hard to obtain supervised training data for the video decomposition problem, we adopt and
extend the approach of~\cite{alayrac19centrifuge} and synthetically generate the training data. This by construction provides direct access to one meaningful ground truth decomposition.
Specifically, we start from two real videos $V_{1},V_{2}\in\mathbb{R}^{T\times W \times H \times 3}$.
These videos are mixed together to generate a training video $V\in\mathbb{R}^{T\times W \times H \times 3}$:
\begin{equation}
\label{eq:generation}
V = \bm{\alpha} \odot V_{1} + (\bm{1}-\bm{\alpha})\odot V_{2},
\end{equation}
where $\bm{\alpha}\in\left[0,1\right]^{T\times W \times H}$ is 
a composing mask.

We explore two ways to generate the composing mask $\bm{\alpha}$. 
The first one is \emph{transparent blending}, used by~\cite{alayrac19centrifuge}, where $\bm{\alpha}=\frac{1}{2}\bm{1}$.
While attractive because of its simplicity, it does not capture the full complexity of the  real world compositions we wish to address, such as occlusions. 
For this reason, we also explore a second strategy, referred to as \emph{occlusion blending}, where $\bm{\alpha}$ is allowed to vary in space and takes values $0$ or $1$. 
In more detail, we follow the procedure of \cite{Doersch19} where
spatio-temporal SLIC superpixels \cite{Achanta12} are extracted from $V_{1}$,
and one is chosen at random. The compositing mask $\bm{\alpha}$ is set to
1 inside the superpixel and 0 elsewhere; this produces mixtures
of completely transparent or completely opaque spatio-temporal regions.
The impact of the $\bm{\alpha}$ sampling strategy on the final performance is explored in Section~\ref{subsec:quant}.

\vspace{3mm}
\noindent
\textbf{Training loss: without control.}
By construction, for an input training video $V$ we know that one valid decomposition is into $V_1$ and $V_2$.
However, when training \emph{without control}, there is no easy way to know beforehand the order in which output videos are produced by the network.
We therefore optimize the network weights to minimize the following permutation invariant reconstruction loss~\cite{alayrac19centrifuge}:
\begin{equation}
\label{eq:pil}
\mathcal{L}_{\text{pil}}\left(\{V_{1}, V_{2}\}, \bm{O}\right)=\min_{(i,j) | i\neq j} \ell(V_{1}, O_i)+\ell(V_{2}, O_j),
\end{equation}
where $\ell$ is a video reconstruction loss, \eg a pixel wise error loss (see Section~\ref{sec:exp} for our particular choice).

\begin{figure}[t]
	\begin{center}
		\includegraphics[width=\linewidth]{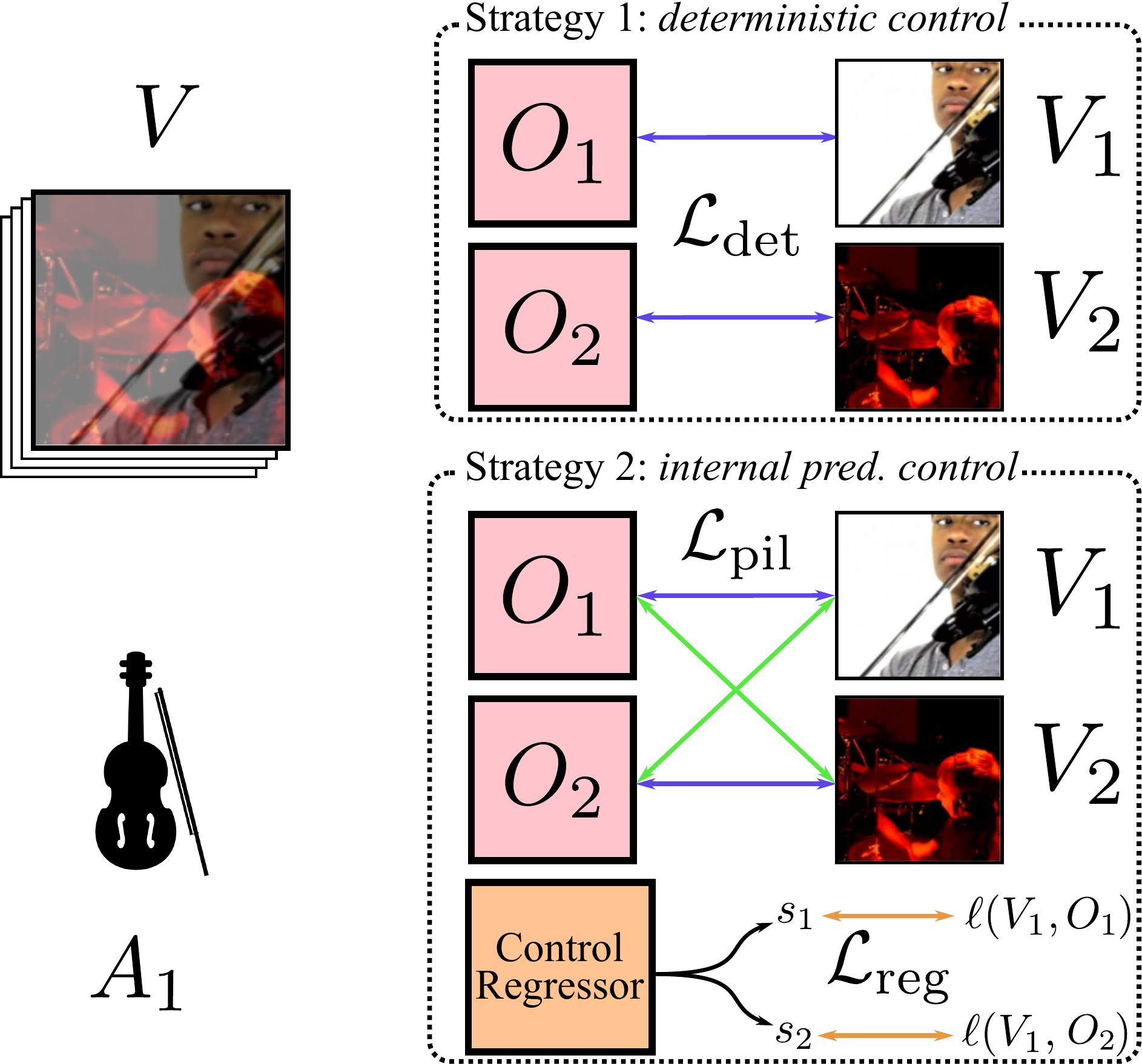}
	\end{center}
	\vspace{-0.4cm}
	\caption{\small {\bf Audio control strategies for video decomposition.}
In this example, the inputs are the video $V$, a composition of $V_1$ showing a violin and $V_2$ showing drums, and an audio control signal, $A_1$, being the sound of the violin.
With \emph{deterministic control}, $V_1$ is forced to be put in output slot $O_1$ (and therefore $V_2$ in $O_2$).
With \emph{internal prediction control}, the network can freely order the output videos, so is trained with the permutation invariant loss, but it contains an additional control regressor module which is trained to point to the desired output.
 \label{fig:audio}}
 \vspace{-0.3cm}
\end{figure}

\vspace{3mm}
\noindent
\textbf{Training loss: with control.}
When training with audio as the control signal, the audio of one video
($V_1$ without loss of generality) is also provided.
This potentially removes the need for the permutation invariant loss required in the no-control case, but the loss depends on the choice of control strategy.
The two proposed strategies are illustrated in Figure~\ref{fig:audio} and described next.

\vspace{3mm}
\noindent
\emph{Deterministic control loss.}
Here, the network is forced to output the desired video $V_1$ as $O_1$ so a natural loss is:
\begin{equation}
\label{eq:det}
\mathcal{L}_{\text{det}}\left(\{V_{1}, V_{2}\}, \bm{O}\right)= \ell(V_{1}, O_1)+\ell(V_{2}, O_2).
\end{equation}
Note that for this loss the number of output videos has to be restricted to $n=2$.
This limitation is another drawback of \emph{deterministic control} as it allows less freedom to propose multiple output video options.

\noindent
\emph{Internal prediction loss.}
In this strategy, the network freely decomposes the input video into outputs,
and therefore the training loss is the same permutation invariant loss as
for the no-control case \eqref{eq:pil}.
In addition, the network also points
to the output which corresponds to the desired video, 
where the pointing mechanism is implemented as a module
which outputs $n$ real values $\bm{s}=(s_i)_{i=1}^n$, one for each output video.
These represent predicted dissimilarity between the desired video and output videos,
and the attended output is chosen as $\argmin_i{s_i}$.
This module is trained with the following regression loss:
\begin{equation}
\label{eq:reg}
\mathcal{L}_{\text{reg}}\left(V_{1}, \bm{s}\right)= \sum_{i=1}^n |s_i - \ell(V_{1}, \text{\textbf{sg}}(O_i))|,
\end{equation}
where \text{\textbf{sg}} is the \texttt{stop\_gradient} operator.
Stopping the gradient flow is important as it ensures that the only effect
of training the module is to learn to point to the desired video.
Its training is not allowed to influence the output videos themselves,
as if it did, it could sacrifice the reconstruction quality in order to set
an easier regression problem for itself.

\section{Experiments}
\label{sec:exp}

This section evaluates the merits of the proposed
Compositional Centrifuge ($\texttt{C}^\texttt{2}$) compared to previous work,
performs ablation studies,
investigates attention control via the audio control signal
and the effectiveness of the two proposed attention control strategies
of the Controllable Compositional Centrifuge ($\texttt{C}^\texttt{3}$),
followed by qualitative decomposition examples on natural videos, and evaluation on
the downstream task of action recognition.

\vspace{3mm}
\noindent
\textbf{Implementation details.}
Following \cite{Mathieu2016,alayrac19centrifuge}, in all experiments we use the following video reconstruction loss, defined for videos $U$ and $V$ as:
\begin{equation*}
\ell(U, V) = \frac{1}{2T} \left(\sum_t \Vert U_t-V_t \Vert_1 + \Vert \nabla(U_t)-\nabla(V_t) \Vert_1\right),
\end{equation*}
\noindent where $\Vert \cdot \Vert_1$ is the L1 norm and $\nabla(\cdot)$ is the spatial gradient operator.

All models are trained and evaluated on the blended versions of the training and validation sets of the Kinetics-600 dataset~\cite{Carreira-Kinetics600-2018}.
Training is done using stochastic gradient descent with momentum for 124k iterations, using batch size 128. We employed a learning rate schedule, dividing by 10 the initial learning rate of 0.5 after 80k, 100k and 120k iterations. In all experiments we randomly sampled 64-frame clips at 128x128 resolution by taking random crops from videos whose smaller size being resized to 148 pixels.

\begin{table}[t]
	\centering
	\begin{tabular}{@{}cccc@{}}
		\toprule
		Model                   & Loss (Transp.) & Loss (Occl.) & Size \\ \midrule
		Identity                & 0.364 & 0.362 & --          \\ 
		Centrifuge~\cite{alayrac19centrifuge} & 0.149  & 0.253 & 22.6M      \\
		CentrifugePC~\cite{alayrac19centrifuge} & 0.135  & 0.264 & 45.4M     \\ \hline
		$\texttt{C}^\texttt{2}$ w/o masking	& 0.131      &  0.200   & 23.4M \\ 
		$\texttt{C}^\texttt{2}$ & \textbf{0.120}       & \textbf{0.190}  & 27.1M \\ \bottomrule
	\end{tabular}
	\vspace{-0.2cm}
	\caption{\label{table:baselines} \small Model comparison in terms of
		average validation loss for synthetically generated videos with transp(arency)
		and occl(usions), as well as size in millions of parameters.
		All the results are obtained using models with $n=4$ output layers.
		\emph{CentrifugePC} is the predictor-corrector centrifuge~\cite{alayrac19centrifuge},
		\emph{Identity} is a baseline where the output videos are just copies of the input.}
\end{table}

\subsection{Quantitative analysis}
\label{subsec:quant}

In this section, we evaluate the effectiveness of our approaches
through quantitative comparisons on synthetically generated data using
blended versions of the Kinetics-600 videos.

\begin{figure}[t]
	\begin{center}
		\includegraphics[width=\linewidth]{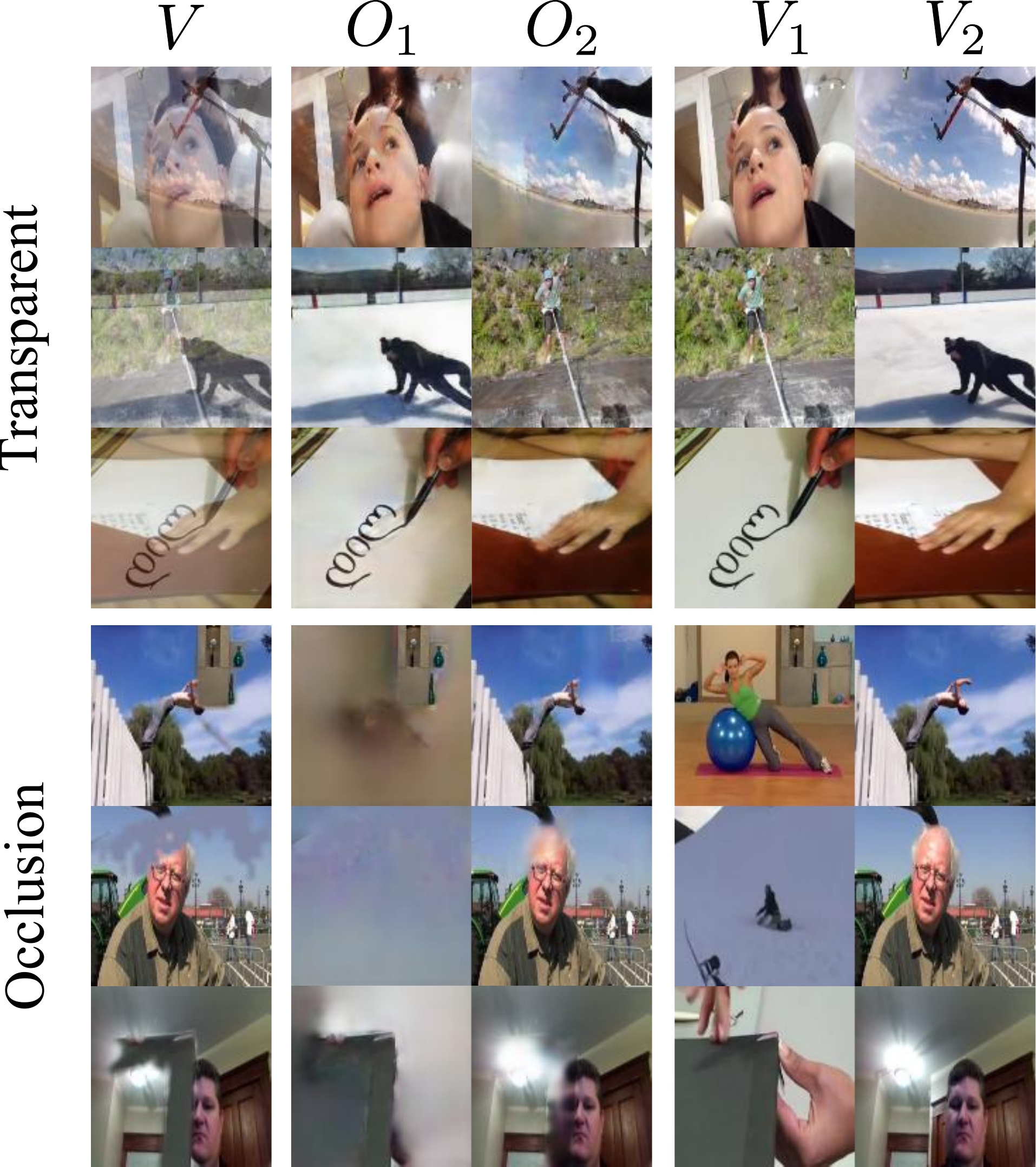}
	\end{center}
	\vspace{-0.4cm}
	\caption{\label{fig:input_mode} \small 
		{\bf Outputs of $\texttt{C}^\texttt{2}$ on blended Kinetics validation clips.}
		Each row shows one example via a representative frame,
		with columns showing the input blended clip $V$, two output videos $O_1$ and $O_2$,
		and the two ground truth clips $V_1$ and $V_2$.
		Top three rows show the network is able to successfully decompose videos with transparencies.
		Bottom three rows show synthetic occlusions -- this is a much harder task where,
		apart from having to detect the occlusions, the network also has to inpaint
		the occluded parts of each video.
		$\texttt{C}^\texttt{2}$ performs satisfactory in such a challenging scenario.
	}
	\vspace{-0.4cm}
\end{figure}

\vspace{3mm}
\noindent \textbf{Effectiveness of the $\texttt{C}^\texttt{2}$ architecture for video decomposition.}
The baseline visual centrifuge achieves a slightly better performance (lower loss) than originally reported~\cite{alayrac19centrifuge}
by training on clips which are twice as long (64 vs 32 frames).
As can be seen in Table~\ref{table:baselines},
our proposed architecture outperforms both the \textit{Centrifuge} baseline~\cite{alayrac19centrifuge},
as well as the twice as large predictor-corrector model of~\cite{alayrac19centrifuge}.
Furthermore, both of our architectural improvements --
the masking and the composition module -- improve the performance (recall that the
baseline \textit{Centrifuge} is equivalent to $\texttt{C}^\texttt{2}$ without the
two improvements). The improvements are especially apparent for
\emph{occlusion blending} since our architecture is explicitly designed
to account for more complicated real-world blending than the simple
\emph{transparency blending} used in~\cite{alayrac19centrifuge}.

\begin{table}[t]
	\centering
	\begin{tabular}{@{}ccc@{}}
		\toprule
		Model  & Loss (Transp.) & Control Acc. \\ \midrule
		$\texttt{C}^\texttt{2}$         & 0.120    & 50\% (chance)     \\
		$\texttt{C}^\texttt{3}$ w/ deterministic control     & 0.191   & 79.1\%     \\
		$\texttt{C}^\texttt{3}$ w/ internal prediction    & 0.119       &  77.7\%    \\ \bottomrule
	\end{tabular}
	\vspace{-0.2cm}
	\caption{\label{table:control_audio} \small Model comparison on average validation reconstruction loss and control accuracy. The controllable models, $\texttt{C}^\texttt{3}$, use audio as the control signal.}
	\vspace{-0.4cm}
\end{table}

\begin{figure*}[t]
	\begin{center}
		\includegraphics[width=\linewidth]{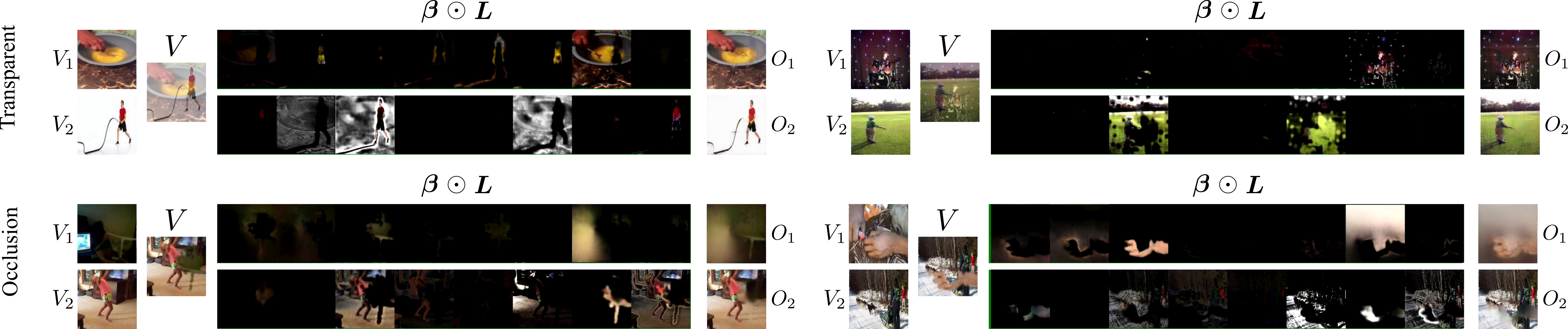}
	\end{center}
	\vspace{-0.4cm}
	\caption{\label{fig:qual_compos} \small 
		{\bf Visualization of the internals of the compositional model.}
		Recall that the $\texttt{C}^\texttt{2}$ model produces the output videos
		via the \emph{composition module} (Figure~\ref{fig:architectures_comp})
		which multiplies the layers $\bm{L}$ with composing coefficients $\bm{\beta}$.
		Here we visualize the individual $\bm{\beta} \odot \bm{L}$ terms which
		when added together form the output videos.
		It can be observed that the layers and composing coefficient indeed
		decompose the input video $V$ into its constituent parts, for both the
		transparent and occlusion blending.
	}
	\vspace{-0.5cm}
\end{figure*}

\vspace{3mm}
\noindent \textbf{Attention control.}
The effectiveness of the two proposed attention control strategies
using the audio control signal is evaluated next.
Apart from comparing the reconstruction quality, we also contrast the methods
in terms of their \emph{control accuracy}, \ie their ability to output the
desired video into the correct output slot.
For a given video $V$ (composed of videos $V_1$ and $V_2$) and audio control signal $A_1$,
the output is deemed to be correctly controlled if the chosen output slot $O_c$
reconstructs the desired video $V_1$ well. Recall that the `chosen output slot'
is simply slot $O_c=O_1$ for the \emph{deterministic control}, and predicted
by the \emph{control regressor} as $O_{\argmin_i(s_i)}$
for the \emph{internal prediction control}.
The chosen output video $O_c$ is deemed to reconstruct the desired video well if
its reconstruction loss is the smallest out of all outputs (up to a threshold $t=0.2*(\max_i \ell(V_{1}, O_i) -\min_i \ell(V_{1}, O_i))$ to account for potentially nearly identical outputs when outputing more than 2 layers):
$\ell(V_{1}, O_c) < \min_i \ell(V_{1}, O_i) + t$.

Table~\ref{table:control_audio} evaluates control performance across different models with the \textit{transparency blending}.
It shows that the non-controllable
$\texttt{C}^\texttt{2}$ network, as expected, achieves control accuracy equal to
random chance, while the two controllable variants of $\texttt{C}^\texttt{3}$
indeed exhibit highly controllable behaviour.
The two strategies are comparable on control accuracy, while 
\emph{internal prediction control} clearly beats \emph{deterministic control}
in terms of reconstruction loss,
confirming our intuition that \emph{deterministic control} imposes overly
tight constraints on the network.

\begin{figure}[t]
	\begin{center}
		\includegraphics[width=\linewidth]{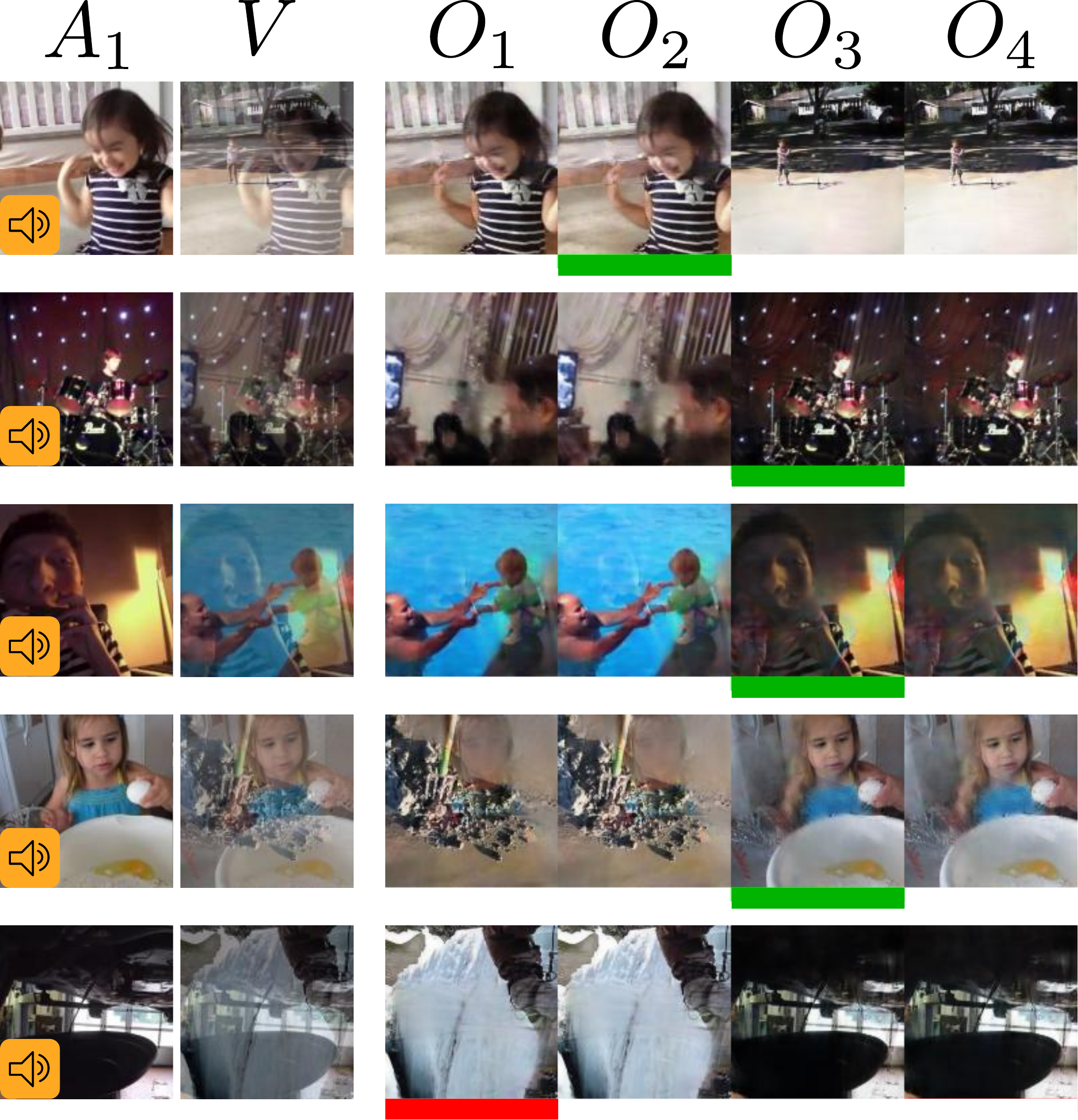}
	\end{center}
\vspace{-0.4cm}
	\caption{\label{fig:qual_control} \small 
		{\bf Qualitative results of $\texttt{C}^\texttt{3}$ with internal prediction.}
		For visualization purposes, as it is hard to display sound, we show a frame of the video from which we use the \emph{audio} as control on the left most column ($A_1$). $V$ (second column) represents the visual input to the model. The right 4 columns are the outputs of $\texttt{C}^\texttt{3}$. 
		All examples exhibit good reconstruction error.
		The first four rows illustrate accurate control behaviour, where $\texttt{C}^\texttt{3}$ has correctly predicted the output that corresponds to the \emph{control} signal (illustrated by a green marker under the frame).
		The last row illustrates an incorrect control (specified with a red marker under the wrongly chosen frame), where 
		$\texttt{C}^\texttt{3}$ was fooled by a liquid sound that is plausible in the two scenarios.		
		}
		\vspace{-0.4cm}
\end{figure}

\subsection{Qualitative analysis}
\label{subsec:qual}

Here we perform qualitative analysis of the performance of our decomposition networks
and investigate the internal layered representations.

Figure~\ref{fig:input_mode} shows the video decompositions obtained from our
$\texttt{C}^\texttt{2}$ network for transparent and occlusion blending.
The network is able to almost perfectly decompose the videos with transparencies,
while it does a reasonable job of reconstructing videos in the much harder
case where strong occlusions are present and it needs to inpaint parts of the videos
it has never seen.

The internal representations produced by our \emph{layer generator}, which
are combined in the \emph{composition module} to produce the output videos,
are visualized in Figure~\ref{fig:qual_compos}. Our architecture indeed
biases the model towards learning compositionality as the internal layers
show a high degree of independence and specialize towards reconstructing
one of the two constituent videos.

Finally, Figure~\ref{fig:qual_control} shows qualitative results for the
best controllable network, $\texttt{C}^\texttt{3}$ with internal prediction,
where audio is used as the control signal. The network is able to accurately
predict which output slot corresponds to the desired video, making few
mistakes which are often reasonable due to the inherent noisiness and ambiguity
in the sound.

\subsection{Downstream tasks}
\label{subsec:downstream}
In the following, we investigate the usefulness of layered video decomposition
as a preprocessing step for other downstream tasks.

\noindent
\textbf{Graphics.}
Layered video decomposition can be used in various graphics applications,
such as removal of reflections, specularities, shadows, \etc.
Figure~\ref{fig:comparison_testbed} shows some examples of decompositions
of real videos. Compared with previous work of~\cite{alayrac19centrifuge},
as expected from the quantitative results, the decompositions are better
as the produced output videos are more pure.

\noindent
\textbf{Action recognition.}
A natural use case for video decomposition is action recognition in challenging
scenarios with transparencies, reflections and occlusions. Since there are
no action recognition datasets focused on such difficult settings, we again
resort to using blended videos.
A pre-trained I3D action recognition network~\cite{carreira17quovadis} is used
and its performance is measured when the input is pure unblended video,
blended video, and decomposed videos, where the decomposition is performed
using the best baseline model (predictor-corrector centrifuge, CentrifugePC~\cite{alayrac19centrifuge})
or our Compositional Centrifuge ($\texttt{C}^\texttt{2}$).
For the pure video performance, we report the standard top-1 accuracy.

For \emph{transparency} blended videos, the desired outputs are both ground truth labels of the two constituent videos. Therefore, the models make two predictions and are scored 1, 0.5 and 0
depending on whether both predictions are correct, only one or none is, respectively.
When I3D is applied directly on the blended video, the two predictions are naturally
obtained as the two classes with the largest scores.
For the decomposition models, each of the two output videos contributes their highest scoring prediction.

In the case of \emph{occlusion} blended videos, the desired output is the ground truth
label of $V_2$ because there is not enough signal to reconstruct $V_1$
as the blended video only contains a single superpixel from $V_1$.
When I3D is applied directly on the blended video, the top prediction is used.
The decomposition models tend to consistently reconstruct $V_2$ in one particular
output slot, so we apply the I3D network onto the relevant output and
report the top-1 accuracy.

Table~\ref{table:actionreco} shows that decomposition significantly improves the
action recognition performance, while our $\texttt{C}^\texttt{2}$ strongly outperforms
the baseline CentrifugePC~\cite{alayrac19centrifuge} for both blending strategies.
There is still a gap between $\texttt{C}^\texttt{2}$ and the pure video performance,
but this is understandable as blended videos are much more challenging.

\begin{figure}[t]
	\begin{center}
		\includegraphics[width=\linewidth]{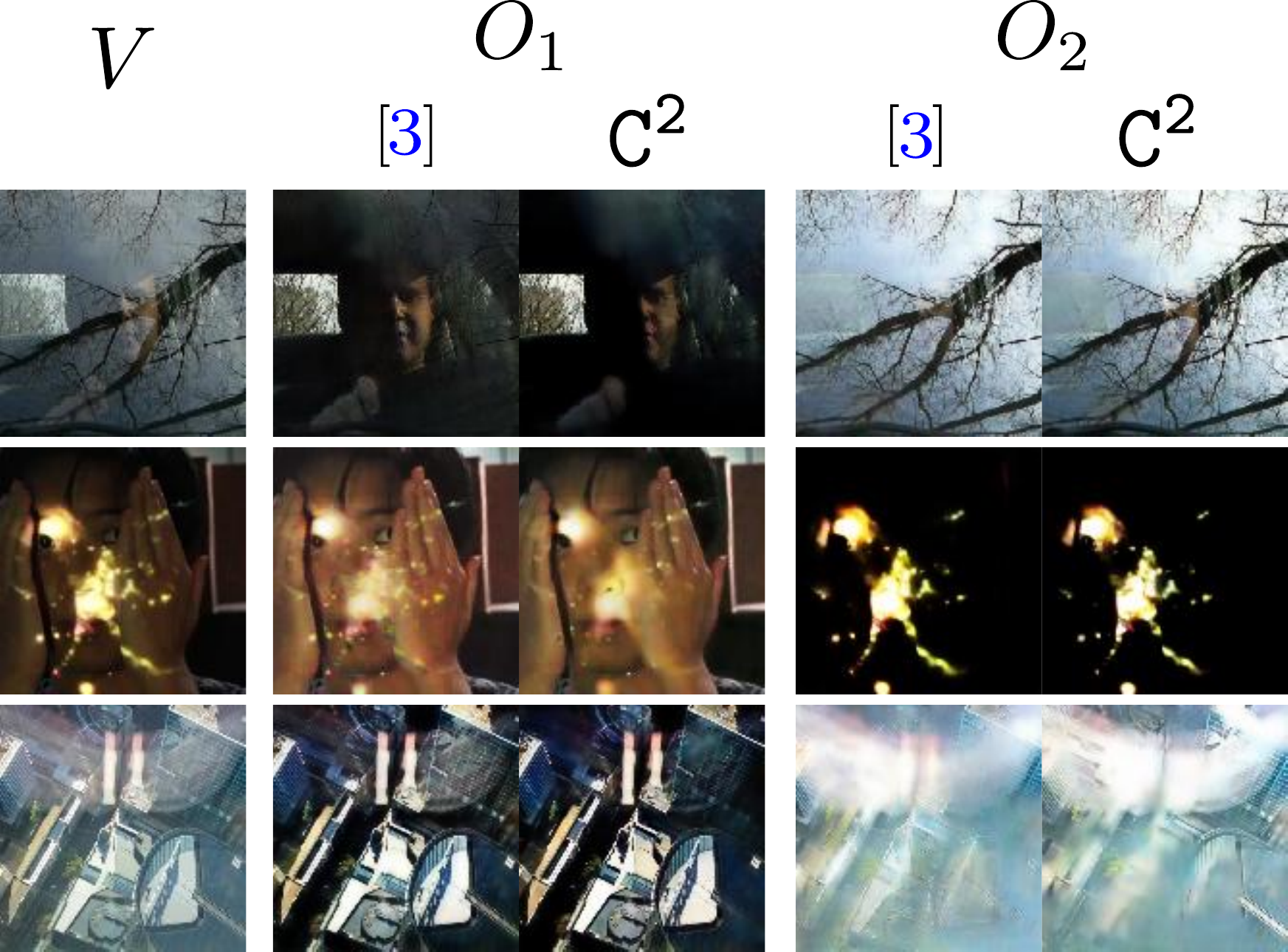}
	\end{center}
	\vspace{-0.4cm}
	\caption{\label{fig:comparison_testbed} \small 
		{\bf Comparison of our $\texttt{C}^\texttt{2}$ model against~\cite{alayrac19centrifuge} on real-world videos.}
The input video is shown on the left, and the output videos of $\texttt{C}^\texttt{2}$
and \cite{alayrac19centrifuge} are interleaved in the remaining columns for easier
comparison. While both models manage to decompose the videos reasonably well,
$\texttt{C}^\texttt{2}$ achieves less leakage of one video into another.
For example, $\texttt{C}^\texttt{2}$ versus \cite{alayrac19centrifuge} output $O_1$
(first row) removes the reflections of branches on the right side better,
(second row) has fewer yellow circles of light,
and
(third row) makes the large circular reflection in
the top half of the image much fainter.
}
\vspace{-0.4cm}
\end{figure}

\begin{table}[t]
	\centering
	\begin{tabular}{@{}ccc@{}}
		\toprule
		Mode                   & Acc. (Transp.) & Acc. (Occl.) \\ \midrule
		I3D -- pure video                & 59.5   & 59.5     \\ \midrule
		I3D 	 		    & 22.1        & 21.3   \\
		CentrifugePC~\cite{alayrac19centrifuge} + I3D & 34.4 & 21.5             \\
		$\texttt{C}^\texttt{2}$ + I3D & 40.1 &   24.7        \\ \bottomrule
	\end{tabular}
	\vspace{-0.2cm}
	\caption{\label{table:actionreco} \small Action recognition accuracy on the Kinetics-600 validation set when the input to a pre-trained I3D classifier is a pure -- non-blended -- video (top row), a blended video directly passed through I3D, or a blended video that is first unblended using a layer decomposition model. The two columns show accuracies for two different blending processes: transparent and occluding.}
	\vspace{-0.4cm}
\end{table}

\section{Conclusion}

General vision systems, that can serve a variety of purposes, will probably require controllable attention mechanisms. There are just too many possible visual narratives to investigate in natural scenes, for a system with finite computational power to pursue them all at once, always. In this paper we proposed a new compositional model for layered video representation and introduced techniques to make the resulting layers selectable via an external control signal -- in this case sound. We showed that the proposed model can better endure automatically generated transparency and especially occlusions, compared to previous work, and that the layers are selected based on sound cues with accuracies of up to 80\% on 
the blended Kinetics dataset.
As future work we would like to train our model on more naturally-looking occlusions, possibly by generating the composing mask using supervised segmentations instead of unsupervised superpixels.

{\small
\bibliographystyle{ieee_fullname}
\bibliography{egbib}
}

\clearpage
\appendix

\begin{figure*}[t]
	\begin{center}
		\includegraphics[width=\linewidth]{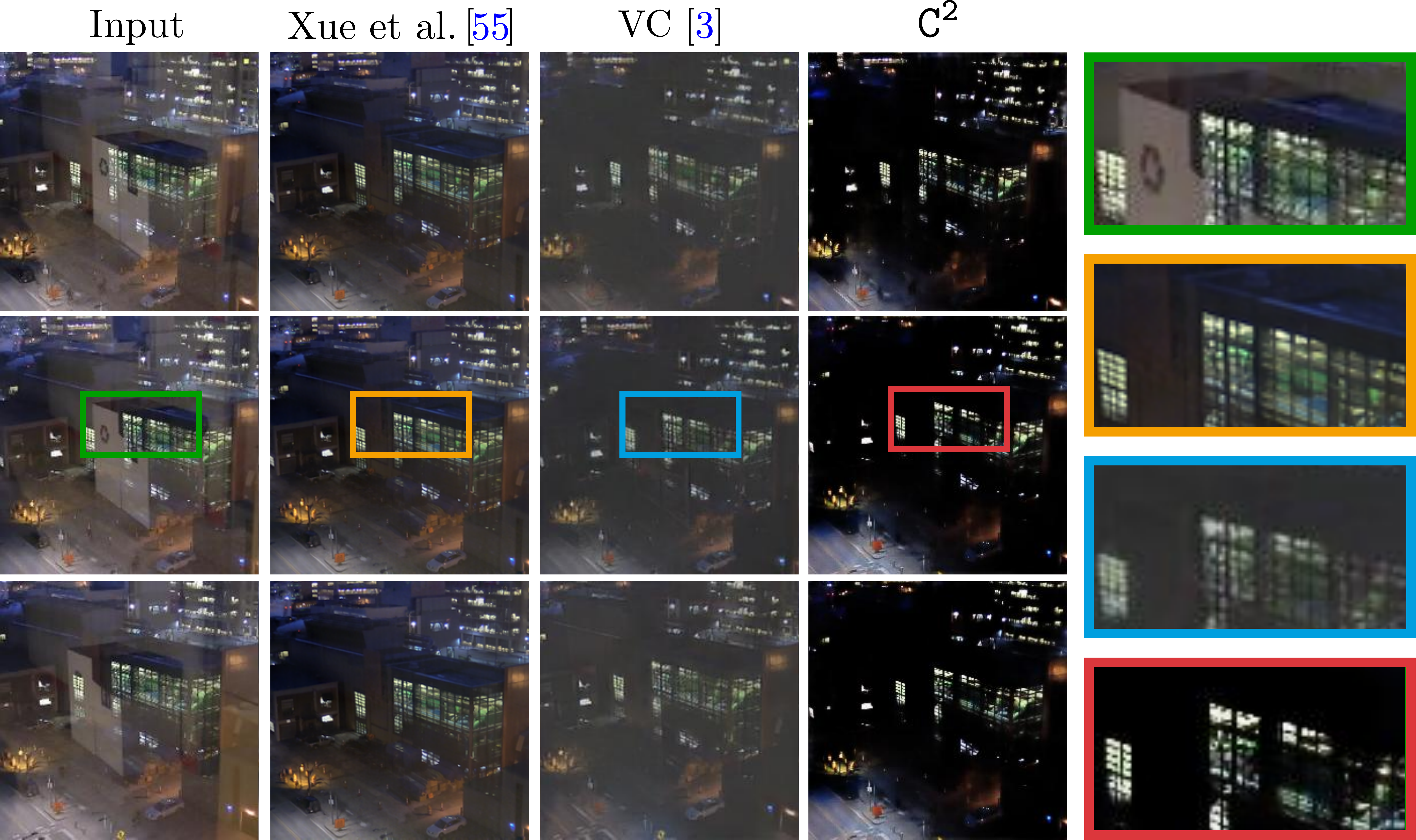}
	\end{center}
	\caption{\label{fig:comparison_freeman} 
		{Qualitative comparison of $\texttt{C}^\texttt{2}$ with other works.
	}}
\end{figure*}

\section*{Overview}

In this appendix, we cover three additional aspects: (i) in Section~\ref{app:comparisons}  we include an additional 
qualitative comparison on 
real videos, for which there was not space in the original manuscript;
(ii) Section~\ref{app:architecture} provides details for the network architecture; and finally 
(iii) in Section~\ref{app:c3} we  study how the network uses the audio for control, by 
perturbing the audio.

\section{Additional comparison of $\texttt{C}^\texttt{2}$ with previous work}
\label{app:comparisons}

We compare to previous work~\cite{Xue2015,alayrac19centrifuge} on the task of reflection removal in Figure~\ref{fig:comparison_freeman}. One of the baselines~\cite{Xue2015} uses geometrical modelling and optimization but under strict assumptions (e.g. rigid motion). The second baseline~\cite{alayrac19centrifuge} is trained on the same data as our model. The proposed model generates a sharp video with little reflection left.

\section{Architecture details}
\label{app:architecture}

Figure~\ref{fig:details_archi_wo_audi} illustrates the architecture employed for $\texttt{C}^\texttt{2}$ while Figure~\ref{fig:details_archi_audi} provides full details about the architecture employed for $\texttt{C}^\texttt{3}$ with \textit{internal prediction} control strategy.

\section{Additional quantitative study for $\texttt{C}^\texttt{3}$}
\label{app:c3}

What aspect of the audio control signal is used
for the controlled decomposition? One hypothesis is that
the network latches onto low-level synchronization cues, so that the desired
output video is identified as the one that is in sync with the audio.
An alternative is that the desired video is the one whose semantic content
matches the audio.

\begin{figure}[t]
	\begin{center}
		\includegraphics[width=\linewidth]{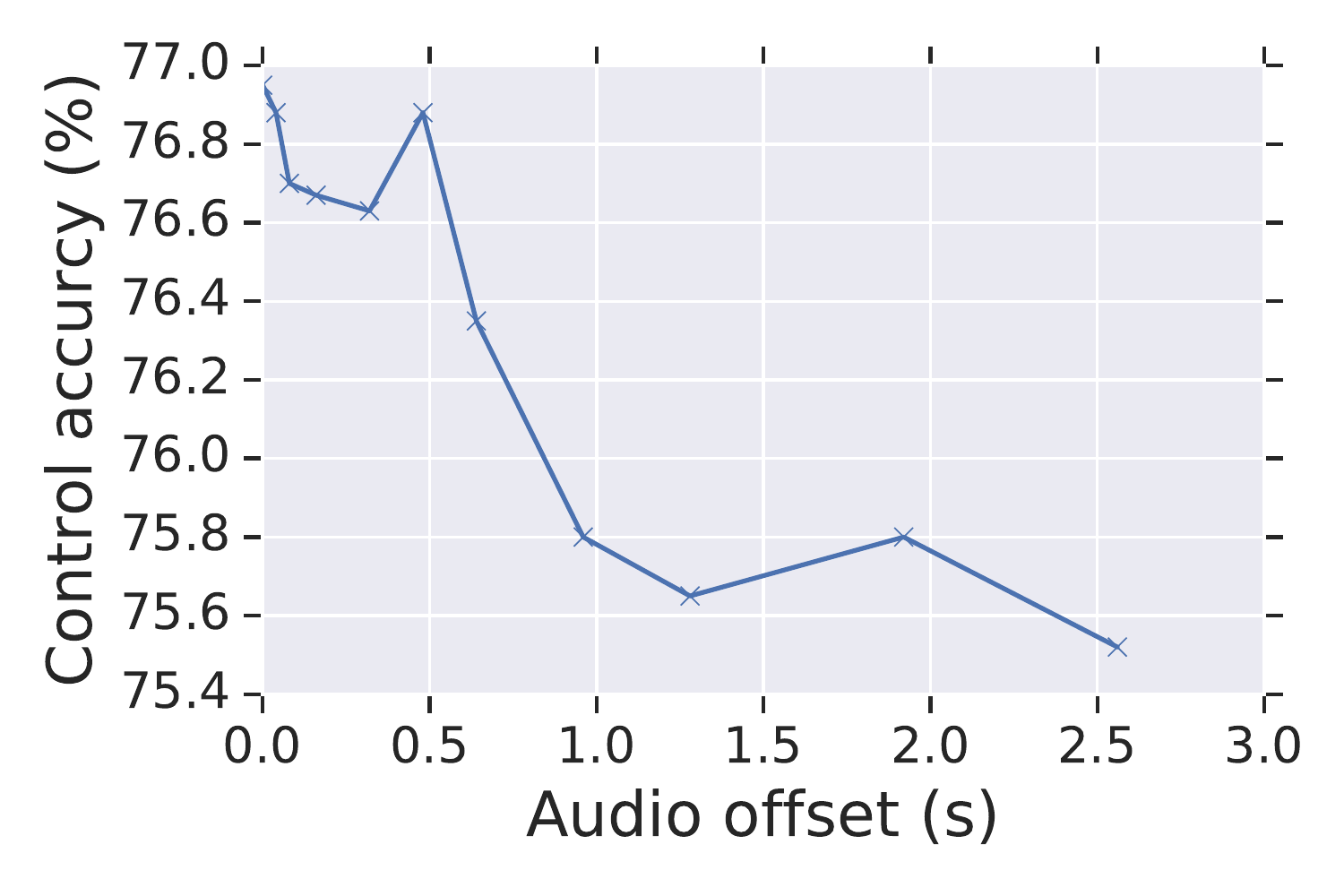}
	\end{center}
	\caption{\label{fig:sync}
		Effect of shifting the control audio signal on the control accuracy.
		Note that the network was trained and tested on 2.56 second clips,
		so a 2.56 second offset corresponds to no overlap between the audio and visual streams.}
\end{figure}

To answer this question, we use the best trained $\texttt{C}^\texttt{3}$ network
with internal prediction control and evaluate its performance
with respect to varying degrees of audio offset.
The experiment is performed on the validation set of Kinetics-600.
Reconstruction loss remains completely unaffected by shifting audio,
while control accuracy deteriorates slightly as the offset is increased,
as shown in Figure~\ref{fig:sync}.
The results suggest that
the network predominantly uses the semantic information contained
in the audio signal as control accuracy only decreases by 1.4 percentage points
with the largest offsets where the audio does not overlap with the visual stream.
However, some synchronization information is probably used as audio offset does
have an adverse effect on control accuracy, and there is a sharp drop
at relatively small offsets of 0.5-1s.
There is scope for exploiting the synchronization signal further
as it might provide a boost in control accuracy. A potential approach
includes using a training curriculum analogous to \cite{Korbar18}.

\begin{figure*}[t]
	\begin{center}
		\includegraphics[width=\linewidth]{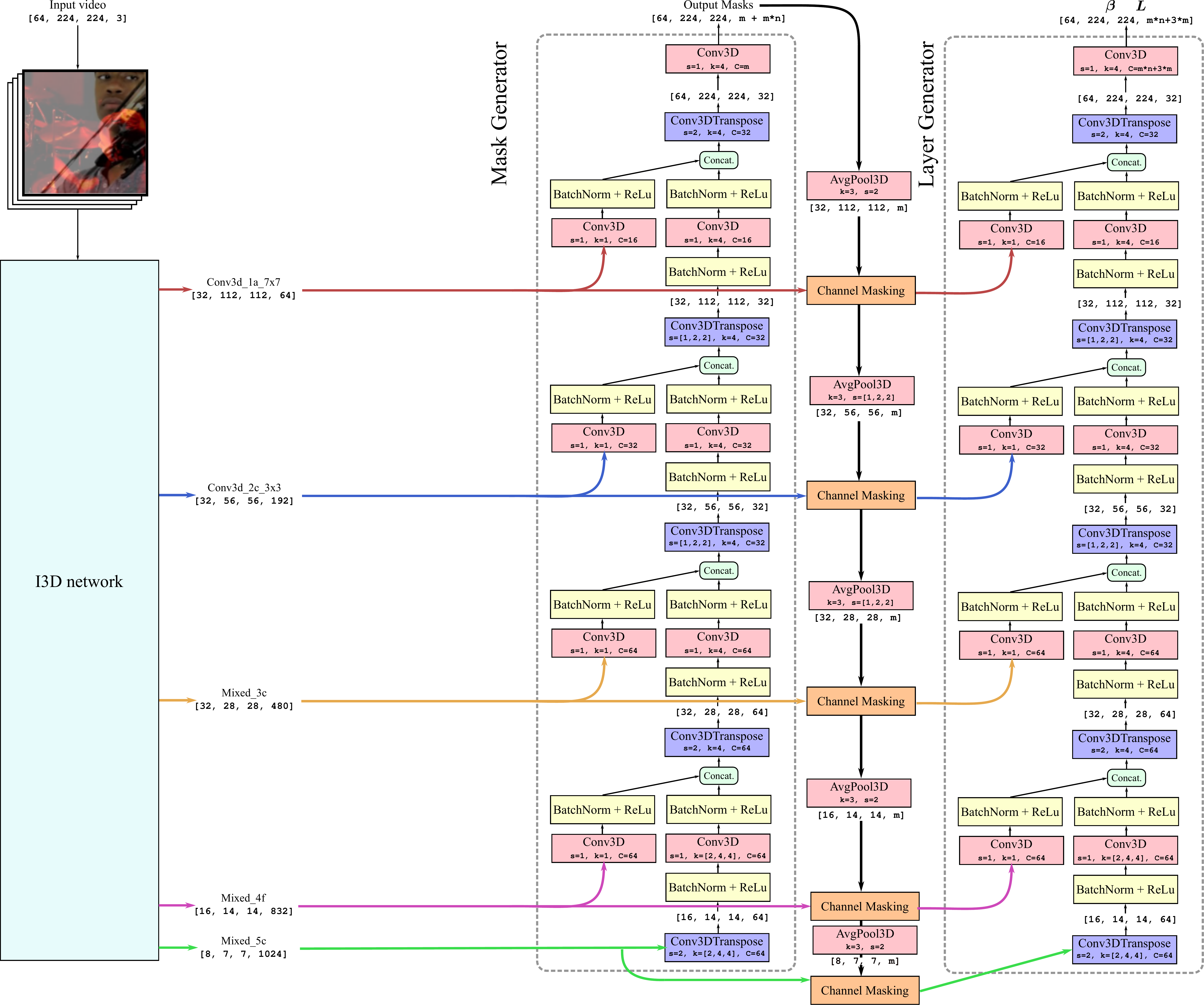}
	\end{center}
	\caption{\label{fig:details_archi_wo_audi} 
		{Details of the architecture used for $\texttt{C}^\texttt{2}$. The `Channel Masking` block corresponds to the masking procedure described in equation~(1) of the main paper. 
	}}
\end{figure*}

\begin{figure*}[t]
	\begin{center}
		\includegraphics[width=\linewidth]{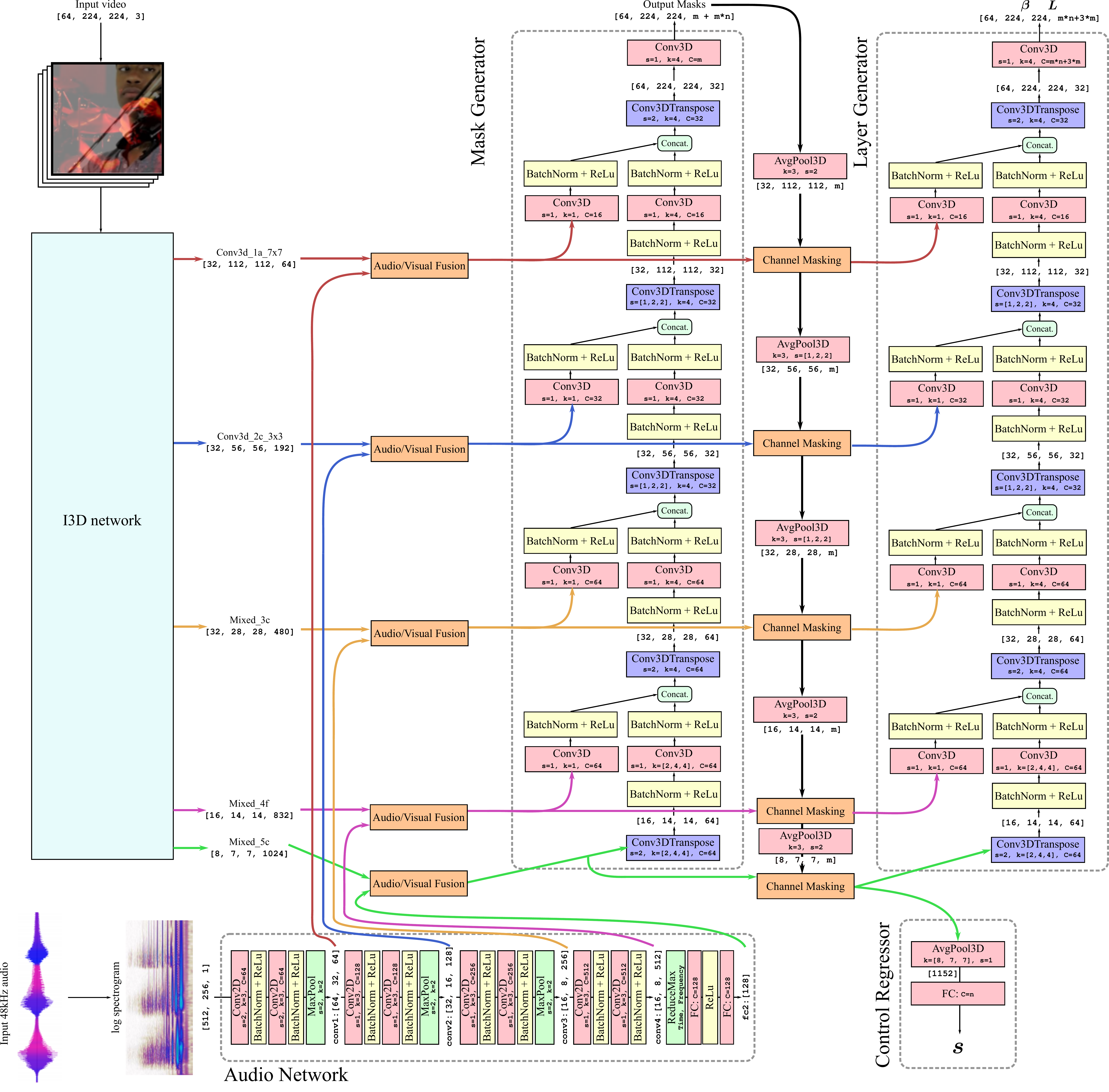}
	\end{center}
	\caption{\label{fig:details_archi_audi} 
		{Details of the architecture used for $\texttt{C}^\texttt{3}$ with \textit{internal prediction} control. The `Channel Masking` block corresponds to the masking procedure described in equation~(1) of the main paper. 
			The `Audio/Visual Fusion` block matches the \textbf{Audio-visual fusion} procedure described in Section~3.2 of the main paper and illustrated in Figure~3.	
	}}
\end{figure*}

\end{document}